\documentclass{article}

\usepackage{microtype}
\usepackage{graphicx}
\usepackage{subfigure}
\usepackage{booktabs} 

\usepackage{hyperref}

\usepackage[accepted]{icml2025}

\usepackage{amsmath}
\usepackage{amssymb}
\usepackage{mathtools}
\usepackage{amsthm}
\usepackage{fontawesome5} 

\usepackage[capitalize,noabbrev]{cleveref}

\theoremstyle{plain}

\theoremstyle{definition}

\theoremstyle{remark}

\usepackage[textsize=tiny]{todonotes}

\usepackage{multirow} 
\icmltitlerunning{GeoPixel: Pixel Grounding Large Multimodal Model in Remote Sensing}

\begin{document}

\twocolumn[

\icmltitle{GeoPixel\texorpdfstring{\includegraphics[width=0.04\textwidth]{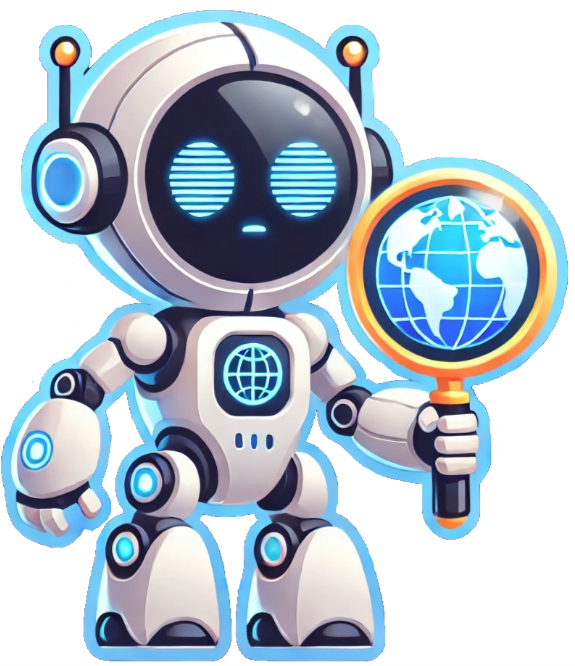}}{:}: Pixel Grounding Large Multimodal Model in Remote Sensing}




\begin{icmlauthorlist}
\icmlauthor{Akashah Shabbir}{MBZUAI}
\icmlauthor{Mohammed Zumri}{MBZUAI}
\icmlauthor{Mohammed Bennamoun}{uwa}
\icmlauthor{Fahad S. Khan}{MBZUAI,sch}
\icmlauthor{Salman Khan}{MBZUAI,ANU}
\end{icmlauthorlist}

\begin{center}
\parbox{0.8\textwidth}{
\begin{center}
{\textsuperscript{1}Mohamed bin Zayed University of AI,\textsuperscript{2}The University of Western Australia, 
\textsuperscript{3}Linköping University, 
\textsuperscript{4}Australian National University} \\
\end{center}
}
\end{center}
\begin{center}
{\tt\small \{akashah.shabbir,mohammed.zumri\}@mbzuai.ac.ae} \\
\end{center}
\begin{center}
\href{https://github.com/mbzuai-oryx/GeoPixel}{\faGlobe\, \tt\small https://github.com/mbzuai-oryx/GeoPixel}
\end{center}

\icmlaffiliation{MBZUAI}{Mohamed bin Zayed University of Artificial Intelligence}
\icmlaffiliation{ANU}{Australian National University}
\icmlaffiliation{sch}{Link\"oping University}
\icmlaffiliation{uwa}{The University of Western Australia}

\icmlcorrespondingauthor{Akashah Shabbir}{akashah.shabbir@mbzuai.ac.ae}
\icmlcorrespondingauthor{Mohammed Zumri}{mohammed.zumri@mbzuai.ac.ae}


\vskip 0.3in
]




\begin{abstract}
\textit{Recent advances in large multimodal models (LMMs) have recognized fine-grained grounding as an imperative factor of visual understanding and dialogue. However, the benefits of such representation in LMMs are limited to the natural image domain, and these models perform poorly for remote sensing (RS). The distinct overhead viewpoint, scale variation, and presence of small objects in high-resolution RS imagery present a unique challenge in region-level comprehension. Moreover, the development of the grounding conversation capability of LMMs within RS is hindered by the lack of granular, RS domain-specific grounded data. Addressing these limitations, we propose GeoPixel - the first end-to-end high-resolution RS-LMM that supports pixel-level grounding. This capability allows fine-grained visual perception by generating interleaved masks in conversation. GeoPixel supports up to 4K HD resolution in any aspect ratio, ideal for high-precision RS image analysis. To support the grounded conversation generation (GCG) in RS imagery, we curate a visually grounded dataset GeoPixelD through a semi-automated pipeline that utilizes set-of-marks prompting and spatial priors tailored for RS data to methodically control the data generation process. GeoPixel demonstrates superior performance in pixel-level comprehension, surpassing existing LMMs in both single-target and multi-target segmentation tasks. Our methodological ablation studies validate the effectiveness of each component in the overall architecture. Our code and data will be publicly released.}
\end{abstract}

\section{Introduction}
\label{intro}

\begin{figure}[t]
\vskip 0.2in
\begin{center}
\centerline{\includegraphics[width=\columnwidth]{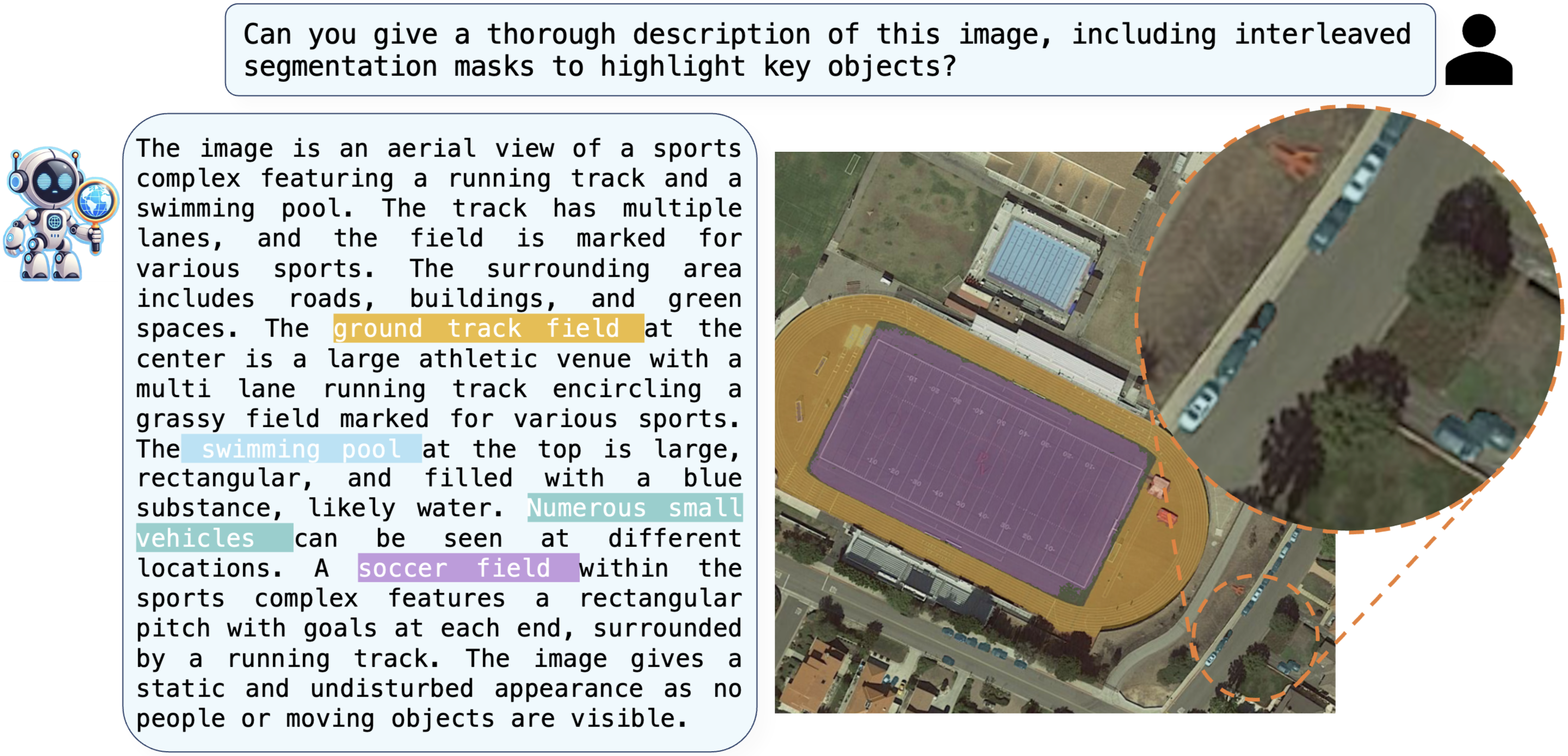}}
\caption{An example of visually grounded detailed descriptions generated by the proposed GeoPixel, highlighting its ability to interpret and segment high-resolution remote sensing imagery with fine-grained precision. The model applies distinct masks to key objects (ground track field, swimming pool, soccer field) and semantic mask to smaller objects (vehicles). It effectively identifies spatial positions (e.g., center, top) and relationships (within the sports complex) while distinguishing between the global context (buildings, roads, green spaces) and localized structures.}
\label{fig:RGC}
\end{center}
\vskip -0.2in
\end{figure}

Recent large multimodal models (LMMs)~\cite{liu2024llavanext,NEURIPS2023_9a6a435e,bai2023qwen,Chen_2024_CVPR_internVL} have utilized the foundational capabilities of Large Language Models (LLMs)~\cite{touvron2023llama,vicuna2023,javaheripi2phi,qwen} and successfully expanded their horizon to the visual modality with promising capabilities. Recent LMMs can not only perform visual recognition, but also excel in advanced perception and reasoning required for vision-language tasks such as visual question answers, image captioning, visual grounding, and referring expression segmentation. Grounding LMMs~\cite{10655326,ma2025groma,zhao2023bubogpt} have further advanced the fine-grained context-aware interpretation of complex visual information by allowing textual outputs to be associated with object instances. Facilitated by large-scale data in the natural images domain, grounding multimodal models pre-trained on extensive datasets have shown impressive capabilities, achieving performance levels comparable to specialist models.

However, with increasing granularity of vision and language understanding, these general domain models exhibit significant limitations in adequately supporting complex earth observation tasks. The performance degradation is influenced not only by the unique vantage point inherent to remote sensing (RS) images but also by large variations in the objects' size and orientation. Moreover, in high-resolution remote sensing imagery, objects of interest may exhibit challenging-to-segment spatial footprints, such as narrow bridges that connect urban landscapes and play a critical role in city traffic planning, adding further complexity to the task.

Existing vision language models in RS~\cite{luo2024skysensegpt,10547418,10655072} use quantized coordinates in the form of bounding boxes to localize and ground objects in their response. Such a representation structure is not adequate to associate correct object semantics and also adds a computational burden to the LLM that scales with the number of distinguishable objects.  Furthermore, monitoring the geospatial environment and its entities demands a broader spatial perspective, now increasingly achievable through advancements in RS technologies that provide high-resolution imagery. However, despite the availability of such rich data, current LMMs in RS struggle to fully exploit this spatial detail. 
These models often struggle with suboptimal resolution capabilities, hindering their ability to capture the intricate patterns present in high-resolution RS images. In addition, existing RS datasets often lack fine-grained spatial association between objects and their corresponding linguistic descriptions.

\begin{table*}[ht]

\centering
\caption{Comparison of remote sensing large multimodal models (RS-LMMs), focusing on their grounding capabilities. The `Region Output' column highlights the model's ability to associate objects with specific spatial regions. Existing models primarily utilize LLMs to generate bounding box coordinates for object grounding. However, none of the current RS-LMMs possess the capability for `pixel grounding', i.e., generating detailed segmentation masks, which are crucial for fine-grained spatial interpretation.}
\label{tab:model_comparison}
\begin{center}
\begin{small}
\begin{sc}
\setlength{\tabcolsep}{2.5pt} 
\resizebox{\textwidth}{!}{
\begin{tabular}{lcccccc}
    \toprule
    \multirow{2}{*}{\small\textbf{Models}} & \multirow{2}{*}{\small\textbf{Resolution}} & \multirow{2}{*}{\small\textbf{Image}}  & \small\textbf{Region} & \small\textbf{Region}& \small\textbf{Pixel} & \small\textbf{End to End} \\
     &  &  & \small\textbf{output} & \small\textbf{Decoder} & \small\textbf{Grounding} & \small\textbf{Model} \\
    \midrule
    RSGPT~\cite{hu2023rsgpt} & 224 × 224 &\textcolor{green!70!black}{$\checkmark$} & \textcolor{red}{$\times$} & \textcolor{red}{$\times$} & \textcolor{red}{$\times$} & \textcolor{green!70!black}{$\checkmark$} \\
    H2RSVLM~\cite{pang2024h2rsvlm} & 336 × 336 & \textcolor{green!70!black}{$\checkmark$} & \textcolor{red}{$\times$} & \textcolor{red}{$\times$} & \textcolor{red}{$\times$} & \textcolor{green!70!black}{$\checkmark$}\\
    
    RS-LLaVA~\cite{bazi2024rs} & 336 × 336 & \textcolor{green!70!black}{$\checkmark$} & \textcolor{red}{$\times$} & \textcolor{red}{$\times$} & \textcolor{red}{$\times$} & \textcolor{green!70!black}{$\checkmark$}\\
    
    GeoChat~\cite{10655072} & 504 × 504 & \textcolor{green!70!black}{$\checkmark$} & \textcolor{green!70!black}{$\checkmark$} & \textcolor{red}{$\times$} & \textcolor{red}{$\times$} &  \textcolor{green!70!black}{$\checkmark$} \\
    
    SkyEyeGPT~\cite{zhan2024skyeyegpt} &  448 × 448& \textcolor{green!70!black}{$\checkmark$} & \textcolor{green!70!black}{$\checkmark$} & \textcolor{red}{$\times$} & \textcolor{red}{$\times$}  & \textcolor{green!70!black}{$\checkmark$} \\
    
    EarthGPT~\cite{zhang2024earthgpt} &-& \textcolor{green!70!black}{$\checkmark$} & \textcolor{green!70!black}{$\checkmark$} & \textcolor{red}{$\times$} & \textcolor{red}{$\times$} & \textcolor{green!70!black}{$\checkmark$} \\
    
    LHRS-Bot ~\cite{muhtar2024lhrs} & 224×224 & \textcolor{green!70!black}{$\checkmark$} & \textcolor{green!70!black}{$\checkmark$} & \textcolor{red}{$\times$} & \textcolor{red}{$\times$} & \textcolor{green!70!black}{$\checkmark$} \\

    SkySenseGPT~\cite{luo2024skysensegpt} & 504 × 504& \textcolor{green!70!black}{$\checkmark$} & \textcolor{green!70!black}{$\checkmark$} & \textcolor{red}{$\times$} &
    \textcolor{red}{$\times$} & \textcolor{green!70!black}{$\checkmark$} \\
    \midrule

   {GeoPixel} & \small{dynamic upto 4k} & {\textcolor{green!70!black}{$\checkmark$}} & {\textcolor{green!70!black}{$\checkmark$}} & {\textcolor{green!70!black}{$\checkmark$}} & {\textcolor{green!70!black}{$\checkmark$}} & {\textcolor{green!70!black}{$\checkmark$}}\\
    \bottomrule
\end{tabular}}
\end{sc}
\end{small}
\end{center}
\vskip -0.1in
\end{table*}

To address these issues, we present GeoPixel, a model that can generate a detailed natural language response for a high-resolution RS image with corresponding geospatial object segmentation masks. Our contributions are as follows:
\begin{itemize}
    \item Our proposed LMM, GeoPixel, is explicitly designed for high-resolution RS image analysis with advanced multi-target pixel grounding capability. Our model adaptively divides the input images into local and global regions, enabling efficient encoding and analysis by accommodating up to 4k resolution. 
    \item We create GeoPixelD, a multi-modal grounded conversation generation (GCG) dataset comprising 53,816 grounded phrases linked to 600,817 object masks,
    specifically tailored for RS image understanding. GeoPixelD offers hierarchically structured annotations, providing rich semantic descriptions that integrate both comprehensive, scene-level contextual information and precise, localized object-level details. Extensively granular annotations are created with segmentation masks through a semi-automated, scalable pipeline that integrates prior-informed visual prompting with state-of-the-art LMMs and ensures quality via rigorous verification and filtering steps. 
    \item We introduce a comprehensive benchmark designed for the systematic evaluation of RS LMMs in fine-grained visual understanding tasks. This benchmark includes 5,427 manually validated pairs of referring expressions and segmentation masks, encompassing 61,384 annotated objects in RS imagery within detailed descriptions having an average length of 647 characters. Our benchmark offers a robust basis for assessing the model's capabilities in interpreting and responding to complex, spatially grounded information.

\end{itemize}

\section{Related Work}
\label{sec:related_work}


\textbf{Large Multimodal Models (LMMs):}
LMMs build on the success of LLMs to acquire vision capabilities.
Pioneer works such as LLaVA~\cite{liu2024visual}, MiniGPT-4~\cite{zhu2023minigpt}, InstructBLIP~\cite{NEURIPS2023_9a6a435e} and mPLUG-Owl~\cite{ye2023mplug} aligned visual features with language representations through a vision language connector, enhanced by instruction tuning to improve multimodal integration. Improving beyond image-level understanding, models such as GPT4RoI~\cite{zhang2023gpt4roi}, InternGPT~\cite{liu2023interngpt} and RegionGPT~\cite{Guo_2024_CVPR} introduce regional understanding by allowing inputs such as points, masks, and bounding boxes. Some models feed image coordinates directly into the language model, while others employ additional feature extraction modules to represent specific image regions' features effectively.

\textbf{Grounding LMMs:}
Region-level comprehension is further expanded by models such as Kosmos-2~\cite{peng2024grounding}, Ferret~\cite{you2023ferret}, Shikra~\cite{chen2023shikra}, Pink~\cite{Xuan_2024_CVPR} and LION~\cite{Chen_2024_CVPR} that allow for the precise location of objects in their outputs based on textual descriptions, a capability known as grounding. These models localize objects on a coarse scale using bounding boxes. 
Recent models ~\cite{lai2024lisa,10655326,Xia_2024_CVPR,Ren_2024_CVPR,Zhang_2024_CVPR,liu2023llava_plus} focus on achieving more fine-grained visual and linguistic semantic alignment, by exploring pixel grounding. LISA~\cite{lai2024lisa}, PixelLM~\cite{Ren_2024_CVPR} and GLaMM~\cite{10655326} incorporate a [SEG] token into the LLM's vocabulary, leveraging its corresponding token embedding as a conditioning input for SAM~\cite{kirillov2023segment} to enable segmentation. Additionally, GSVA~\cite{Xia_2024_CVPR} introduces a [REJ] token to explicitly learn to reject specified targets. 
Whereas Llava-plus~\cite{liu2023llava_plus} employs LLMs as agents to assign tasks to the segmentation expert. 

Our work aligns with pixel-grounding approaches, such as those in ~\cite{lai2024lisa, Ren_2024_CVPR, 10655326}. However, these models do not interpret the distinct top-down perspective and cannot differentiate complex spatial arrangements of remote sensing (RS) imagery. In addition, the models' restricted input size, typically limited to dimensions such as 224×224, exacerbates this issue by constraining the field of view and spatial perception.

\textbf{High-Resolution Understanding:}
Vision encoders, such as CLIP ViT~\cite{radford2021learning}, are widely utilized for various vision tasks but are typically constrained by low resolution (e.g. 224×224) 
restricting their applicability in high-resolution (HR) scenarios. To address this limitation, some approaches~\cite{dosovitskiy2021an,bai2023qwen,li2023blip} scale positional encodings within the CLIP model through interpolation to accommodate larger input sizes, while others such as CogAgent~\cite{hong2024cogagent} and Vary~\cite{wei2025vary}, employ an additional HR branch.
Models such as Monkey~\cite{li2024monkey}, SPHNIX~\cite{lin2023sphinx}, Llava-Next~\cite{liu2024llavanext}, IXC2.5~\cite{internlmxcomposer2_5}, Textmonkey~\cite{liu2024textmonkey} and Ureader~\cite{ye2023ureader} divide the image into grids, encoding each section independently to enhance performance on HR text-centric tasks.

\textbf{Remote Sensing (RS) LMMs:}
RSGPT~\cite{hu2023rsgpt} is a pioneering RS model that enables natural language conversation and generates detailed captions. This was followed by GeoChat~\cite{10655072} that supports region-specific inputs and visual grounding through oriented bounding box coordinates in its responses. Furthermore, SkyEyeGPT~\cite{zhan2024skyeyegpt} extends its functionality to RS video captioning, while EarthGPT~\cite{zhang2024earthgpt} and EarthDial~\cite{soni2024earthdial} integrate various multisensor RS interpretation tasks within the LMM framework. 

\begin{figure*}[t]
\vskip 0.2in
\begin{center}
\centerline{\includegraphics[width=1\linewidth]{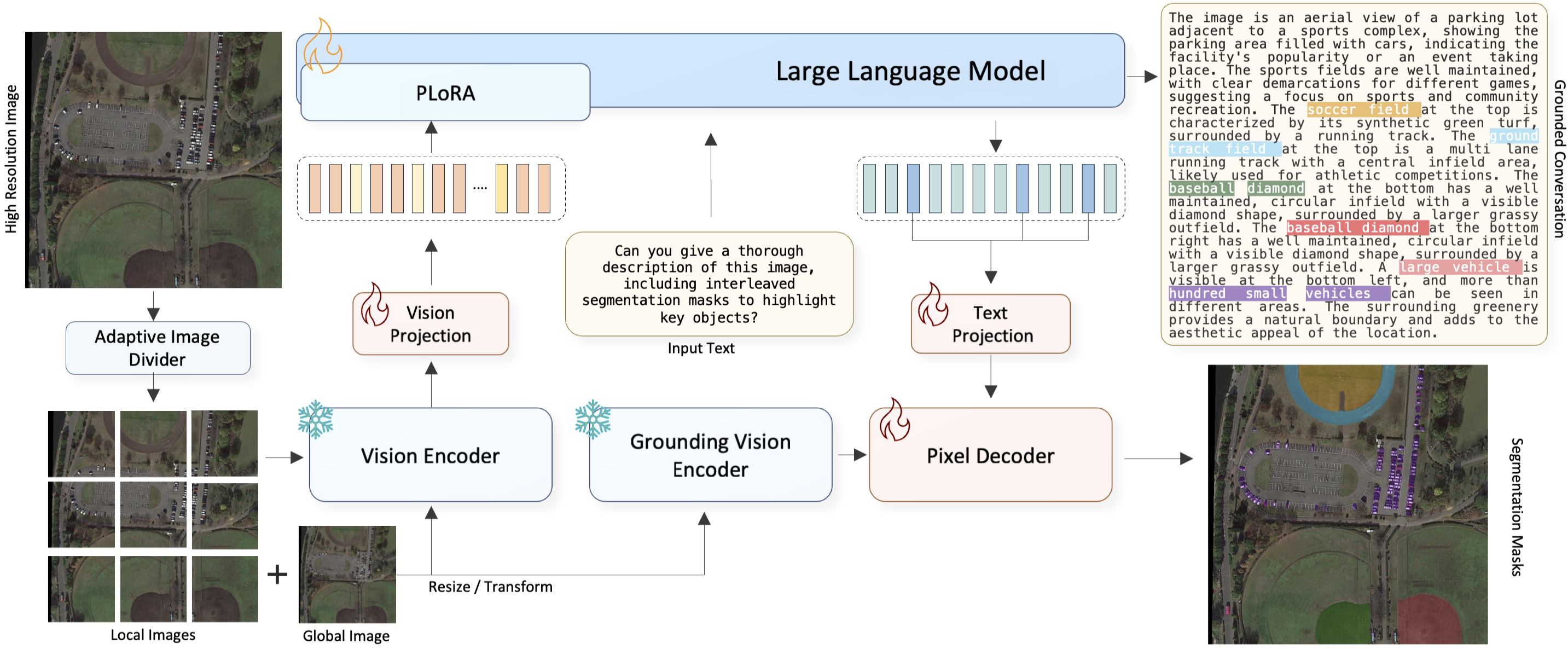}}
\caption{Overview of GeoPixel Architecture: Left: High-resolution RS images are dynamically partitioned into local patches and a resized global view, encoded by a frozen vision encoder. The encodings are projected into the language domain with separator tokens. Middle: Vision tokens, combined with text, are input into the LLM, where pLoRA is applied to vision tokens for efficient and effective multimodal alignment. Right: The corresponding embeddings for the [SEG] tokens are passed to a decoder through text projector, along with vision embeddings from the grounding vision encoders, to generate precise segmentation masks.}
\label{fig:geopixelmodel}
\end{center}
\vskip -0.2in
\end{figure*}

Models such as RS-LLaVA~\cite{bazi2024rs} and H2RSVLM~\cite{pang2024h2rsvlm} improve the interpretation of RS data, with H2RSVLM uniquely recognizing and rejecting unanswerable questions.
SkySenseGPT~\cite{luo2024skysensegpt} contributes by implementing image-level scene graph generation and relation reasoning, while LHRS-Bot~\cite{muhtar2024lhrs} enhances multilevel vision-language alignment.
However, these models operate on low resolution and lack pixel-level understanding and grounding capabilities.

\section{Method}
\label{sec:method}

In the current remote sensing landscape, large multimodal models (LMMs) face significant limitations in terms of grounding and resolution capabilities (as seen in Table~\ref{tab:model_comparison}). Specifically, the outputs generated by these models lack precise spatial and semantic association with the imagery, leading to either ungrounded or only coarsely grounded text. Furthermore, most LMMs operate on relatively low-resolution data, which restricts their ability to perform fine-scale analysis essential for RS tasks such as detailed land use and transportation network extraction, infrastructure mapping, damage assessment, and environmental monitoring. To address these limitations, we present GeoPixel, a model designed to interpret high-resolution remote sensing images and generate finely detailed, pixel-grounded outputs that encompass multiple target objects.

\subsection{GeoPixel Architecture Overview}
GeoPixel primarily consists of 5 components (see Figure~\ref{fig:geopixelmodel}). (1) Adaptive Image Divider (2) Vision Encoder (3) Large Language Model (4) Grounding Vision Encoder (5) Pixel Decoder. The first three components are discussed in Section \ref{subsec:high_res_understanding}, while the latter two in Section \ref{subsec:pixel_grounding}. Jointly, these modules enable high-resolution perception, fine-grained interpretation, and grounding, as detailed below.

\subsection{High Resolution Understanding}
\label{subsec:high_res_understanding}

For high resolution, we adopt the dynamic image partitioning strategy of IXC-2.5~\cite{internlmxcomposer2_5}. Initially, the adaptive image divider processes the input image $x_{img}$, with dimensions $[h_i \times w_i]$, by up-scaling and padding it to align with the closest grid size denoted as $[g_h \times g_w]$.

\begin{equation}
g_h = k_1 \times \mathcal{B}, \quad g_w = k_2 \times \mathcal{B},
\end{equation}
\begin{equation}
\text{s.t., } k_1, k_2 \in \mathbb{N}, \quad k_1 \times k_2 \leq \mathcal{P} \notag
\end{equation}

where $\mathcal{B}$ is the base resolution of the vision encoder and $\mathcal{P}$ is the number of maximum allowable image patches. Subsequently, the image is divided into $k_1 \times k_2$ non-overlapping patches $x_{p_{i,j}}$, where $p = 0, 1, 2, \dots, (k_1 \times k_2 - 1)$, and \( i, j \) denote the row and column indices of each patch in the grid.

\begin{figure*}[t]
 \vskip 0.2in
\begin{center}
\includegraphics[width=1\linewidth]{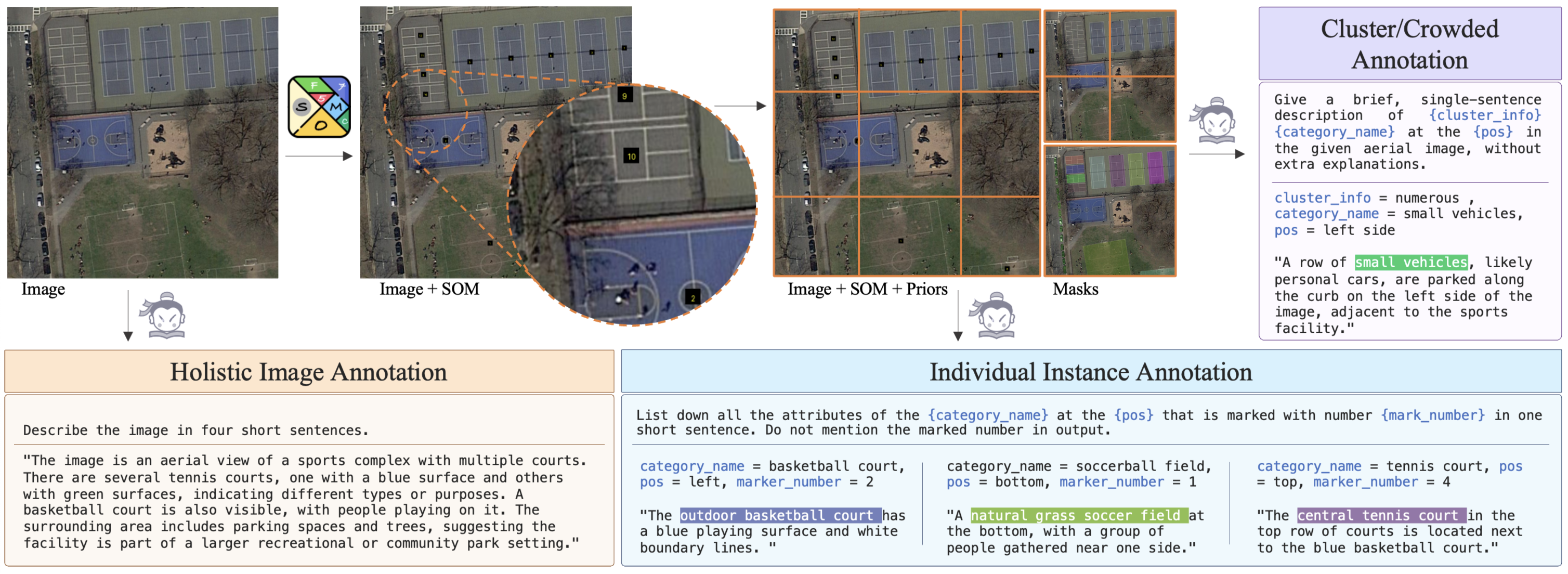} \vspace{-1em}
\caption{The GeoPixelD Annotation Pipeline provides detailed multi-tier descriptions of remote sensing imagery with object phrases aligned precisely with manually annotated masks. It begins with Holistic Image Annotation (bottom left), where an LMM generates concise scene descriptions. Individual Instance Annotation (bottom right) uses spatial(\{pos\}) and categorical (\{catagorory\_name\}) priors with SOM (\{mark\_number\}) prompting to describe key objects. Cluster Annotation (top right) organizes smaller or dense objects using refined grids for precise spatial analysis.}
\label{fig:geopixeld}
\end{center}
\vskip -0.2in
\end{figure*} 

We employ the scaled CLIP ViT-L/14~\cite{internlmxcomposer2_5} as our vision encoder ($\mathcal{I}$), with a base resolution of $\mathcal{B} = 560$, facilitating large patches for enhanced visual representation. Furthermore, a global view $x_{glob}$ is generated by resizing $x_{img}$ to a fixed dimension of $ 560 \times 560 $, aligned with the base resolution $\mathcal{B}$. Feature embeddings of patches $f_{p_{i,j}}$ are appended with a learnable token at the end of each row before flattening and merging~\cite{internlmxcomposer2_4khd}. Finally, global features $f_{glob}$ and patch features $f_p$ are concatenated ($||$) with a special separator ($s_{g}$) inserted between them \cite{ding2019towards}, effectively integrating global semantics with fine-grained local details.
\begin{equation}
x_{v} = \mathcal{P}_v(f_{glob} || s_{g} || f_p)
\end{equation}
\begin{equation}
\text{s.t. } f_{glob} = \mathcal{I}(x_{glob}),\, f_{p_{i,j}} = \mathcal{I}(x_{p_{i,j}}) \notag
\end{equation}

We project the final unified image features onto the LLM, InternLM2 7B model~\cite{cai2024internlm2}, denoted as $\mathcal{L}$, through a two-layer MLP as a vision projector $\mathcal{P}_v$.  InternLM2 is a LLM designed to process sequences of text tokens, where its input consists of a sequence of discrete embeddings derived from textual data. These embeddings correspond to either natural language tokens or special placeholders inserted to represent external modalities.
The placeholder \verb|<IMAGE>| in the input text query $x_{t}$ is a special token that represents the position of the image within the input sequence. When processing multimodal input, this placeholder is replaced with visual features $x_{v}$, extracted from the image, and projected into the same embedding space using $\mathcal{P}_v$.

Partial Low-Rank Adaptation (LoRA)~\cite{internlmxcomposer2} is then applied to ensure efficient alignment of the vision tokens. Partial LoRA is a modality-specific plug-in module designed to align features from a new modality with LLM, preserving the model's inherent capabilities while enriching it with modality-specific insights. By applying low-rank adaptations selectively to visual tokens, Partial LoRA enhances alignment efficiency while reducing the computational cost. Formally, it introduces low-rank matrices $W_A \in \mathbb{R}^{C_r \times C_{in}} $ and $ W_B \in \mathbb{R}^{C_{out} \times C_r} $ within each LLM linear layer, modifying the visual token outputs $ {x}_v $ without altering the language token outputs $ {x}_t $, thus achieving tailored cross-modal integration. 

\subsection{Pixel Grounding}
\label{subsec:pixel_grounding}
To establish grounding in LMM, we initialize the grounding vision encoder ($\mathcal{I}_g$) with a pre-trained SAM-2~\cite{ravi2024sam2} encoder together with a dedicated pixel decoder module ($\mathcal{D}$). The SAM2 visual encoder is a Masked Autoencoder (MAE)~\cite{MAE} pre-trained Hiera~\cite{ryali2023hiera} image encoder having a hierarchical structure that allows the use of multiscale features during decoding. The tokenizer's vocabulary is expanded by incorporating an additional \verb|<SEG>| token, with its corresponding last-layer embedding ($E$) mapped to the decoder through a text projection layer $\mathcal{P}_t$. The text projection is a two-layer MLP that receives embeddings of dimension 4096 and transforms them into the input space of the pixel decoder, which has a dimensionality of 256. 

The pixel decoder processes the image features from the frozen grounding vision encoder, along with projected LLM embeddings, to generate segmentation masks ($M$). The grounding vision encoder (SAM-2) is already pre-trained on large-scale datasets, making it highly effective at extracting robust, generalized image features for segmentation. Freezing the encoder ensures that these pretrained features are preserved. However, the light-weight pixel decoder and projection layer are trained to adapt pretrained vision features for segmentation tasks in GeoPixel.

\begin{equation}
M = \mathcal{D}[\mathcal{I}_g(x_{img}),\mathcal{P}_t(E)]
\end{equation}

Given the variable length of the input image tokens, resulting from adaptive image partitioning, the output embedding mask for \verb|<SEG>| tokens is dynamically adjusted to align with these variations. This configuration ensures accurate detection of the \verb|<SEG>| token and its associated embedding.


\section{GeoPixelD-RS Pixel Grounding Dataset}
\label{sec:geopixeld}

Remote sensing imagery captures intricate semantic information and complex inter-object relationships across diverse spatial scales. To enable LMMs to acquire a detailed comprehension ability, it is essential to integrate broad contextual views with object-level distinction. Addressing the current deficit in datasets capable of facilitating a fine-grained understanding of top-down perspectives, we introduce GeoPixelD, a dataset established to provide hierarchical descriptions derived through automated multilevel image analysis. GeoPixelD structures its descriptions at three primary levels: (1) holistic scene representation, (2) individual instance observations, and (3) densely populated object groups annotations (as depicted in Figure.~\ref{fig:geopixeld}).

\subsection{Holistic Image Annotation}

Initially, we generated descriptive captions for RS images using a robust open source model, IXC~\cite{internlmxcomposer2_5} to capture a comprehensive and diverse image details. We chose the IXC model~\cite{internlmxcomposer2_5} based on a 
comparative study conducted with other state-of-the-art vision language models, where IXC consistently outperformed its counterparts in terms of qualitative performance. These open-ended descriptions are constrained to a limited length, integrated in prompts like, \texttt{"<image> Describe the image in four short sentences."} (Figure~\ref{fig:geopixeld} (bottom left)). Thus, redundancy is effectively minimized in subsequent annotations, and the model is driven to provide a holistic, context-rich depiction of each image. 

\subsection{Individual Instance Annotation}
Next, we identify prominent objects for the depiction and employ a technique known as set-of-mark (SOM) prompting~\cite{yang2023setofmark}. This approach involves adding a distinct set of visual markers over specific regions in an image, providing auxiliary information to obtain visually grounded outputs. However, directly employing this method for aerial imagery, which is characterized by expansive views and diverse objects and landscapes within a single frame, leads to challenges, such as the generation of hallucinated markers and incorrectly associated details (see Figure ~\ref{fig:comp}).

To address the challenge of accurate object description in complex RS images, we implemented an enhanced approach to spatially guide the model. We introduce prior knowledge in the query in the form of category name and location along with a marked number to accurately direct the model and create a comprehensive description of the target object.

Specifically, we partition each image into a 3$\times$3 grid (nine quadrants). For each object, we calculate its positional reference by determining the degree of overlap with these quadrants, thereby localizing it within the grid structure. This quadrant-based localization, combined with categorical labels and marked numbers, is then fed as positional and categorical priors into the LMM, enabling it to focus more accurately on the intended object and retrieve relevant details, a process that proves effective given the densely packed and spatially complex nature of RS imagery, where objects often vary in scale, orientation, and proximity.

In addition, we conducted a comprehensive evaluation of various open-source and proprietary models for prior-informed modified SOM prompting applied to RS imagery (see Figure~\ref{fig:compquery}). The analysis also included a comparative assessment of combined versus individual querying approaches. 
ChatGPT~\cite{openai2023chatgpt} demonstrated the ability to generate detailed descriptions while incorporating inferred information, whereas Gemini~\cite{team2023gemini} and InternVL~\cite{Chen_2024_CVPR_internVL} exhibited repetitive output as the number of target objects within the image increased. InternLM-XComposer~\cite{internlmxcomposer2_5} achieved performance comparable to ChatGPT in terms of the proportion of accurate responses generated and diversity in details.


\subsection{Cluster/Crowd Annotation}
Once prominent large objects are identified, marked and annotated, the remaining objects are grouped or identified along with determining their spatial properties, which is obtained by a structured three-stage positional analysis. In the first stage, the image is divided into a 3$\times$3 grid, with each grid cell assigned a unique identifier corresponding to its spatial location. To enhance alignment with human perceptual tendencies, the central region of the grid is given a larger spatial weight. In the second stage, 2$\times$2 gird is considered for more dispersed objects' localization. Similarly, in the third stage, half image as (1$\times$2 and 2$\times$1) grid is considered to assign positional information. This gridding provides a systematic framework for analyzing the location of clusters as well as large groups of objects within the image. An LMM is then used to describe the group attributes given the quantitative information along with the determined positional information. 

\subsection{Unifying Annotations and Language Marking}
For the preprocessed training subset of the iSAID~\cite{waqas2019isaid} dataset (Appendix ~\ref{Appendix_A}), we derive a total of 16,795 holistic image-level annotations, 36,793 instance-specific annotations, and 17,023 group annotations, collectively encompassing 600,817 objects within RS imagery. The annotations were rigorously filtered to eliminate aerial perspective inconsistencies, removing artifacts such as marker identifiers, fore/background references, distance perception, and contextually inconsistent descriptors. 
\begin{table*}[ht] 
\caption{Performance Comparison on RS-GCG task. LISA$\dagger$ and PixelLM$\dagger$ denote the pretrained LISA and PixelLM models adopted for RS-GCG and finetuned on GeoPixelD training data. GLaMM represents the zero-shot performance, whereas GLaMM-FT refers to the pretrained model finetuned on GeoPixelD. GeoPixel outperforms other models across all metrics.}
\label{tab:res}
\vskip 0.1in
\begin{center}
\begin{small}
\begin{sc}

\setlength{\tabcolsep}{3.5pt} 
\begin{tabular}{lcccccccccccc}
\toprule
&&&\multicolumn{3}{c}{\textbf{Uni-Target}} & \multicolumn{3}{c}{\textbf{Multi-Target}} & \multicolumn{3}{c}{\textbf{Overall}} \\
\cmidrule(lr){4-6} \cmidrule(lr){7-9} \cmidrule(lr){10-12}
\textbf{Model} & \textbf{CIDEr} & \textbf{METEOR} & \textbf{AP50} & \textbf{mIoU} & \textbf{Recall} & \textbf{AP50} & \textbf{mIoU} & \textbf{Recall} & \textbf{AP50} & \textbf{mIoU} & \textbf{Recall} \\
\midrule
GLaMM $_{\text{(CVPR'24)}}$  & 0.1 & 5.8 & 1.2 & 18.1 & 14.8 & 0.5 & 16.5 & 6.3 & 0.5  & 16.9 & 7.1 \\
LISA$\dagger$ $_{\text{(CVPR'24)}}$      & 14.6 & 22.3 & 9.5 & 41.7 & 43.1 & 8.3 & 43.1 & 27.5 & 8.5  & 42.7 & 29.0 \\
PixelLM$\dagger$ $_{\text{(CVPR'24)}}$   & 18.3 & 22.5  & 13.5 & 41.2 & 44.0 & 10.4 & 42.9 & 28.1 & 10.5   & 42.4   & 29.6   \\
GLaMM-ft $_{\text{(CVPR'24)}}$ & 15.7 & 23.0 & 18.8 & 44.4 & 48.5 & 12.4 & 47.1 & 31.1 & 12.5 & 46.4 & 32.8 \\
\midrule
GeoPixel  & \textbf{21.6} & \textbf{24.0} & \textbf{25.5} & \textbf{50.8} & \textbf{55.6} & \textbf{18.0} & \textbf{52.9} & \textbf{37.0} & \textbf{19.0} & \textbf{52.3} & \textbf{38.8} \\
\bottomrule

\end{tabular}

\end{sc}
\end{small}
\end{center}
\vskip -0.1in
\end{table*}
The key noun chunk corresponding to the object category in individual- and group-level annotations is tagged with unique identifiers ('phrase-number'), each linked to an instance or semantic mask, a process termed \emph{text marking}. To unify these hierarchical annotations into a coherent description, the marked annotations are then combined with holistic scene representations to form a single descriptive narrative. We employ a Llama-3.1-instruct 8B~\cite{dubey2024llama} LLM to paraphrase concatenated annotations while preserving their semantic integrity (see Figure~\ref{fig:para}). The LLM processes the concatenated text under strict constraints to retain all marked phrases unchanged, ensuring a consistent link to their associated visual masks. The outputs are rigorously evaluated for consistency, and iterative paraphrasing is applied if any marked phrases are not preserved. By adopting this language marking strategy, the GeoPixelD dataset achieves a robust framework to generate high-quality GCG descriptions that are contextually rich and precisely aligned with visual elements.

A similar procedure is followed for the test set GCG descriptions derived from the iSAID validation subset. Each GCG description within this set undergoes meticulous manual curation, an effort that requires approximately 350 man-hours to ensure annotation completeness. The process includes correcting for any omissions, inaccuracies, or partial annotations, including adjustments to object attributes that do not align with the corresponding image, thereby establishing a high-quality evaluation benchmark.

\section{Experiments}
\label{sec:experiments}

Here, we explain the implementation details, present a comparative performance analysis on Remote Sensing Grounded Conversation Generation (RS-GCG) and Referring Remote Sensing Image Segmentation (RRSIS), and include an ablation study to assess the impact of key components.

\subsection{Implementation Details}
The model weights are initialized using the pre-trained InternLM-XComposer-2.5 model (IXC-2.5) with 7B parameters, utilizing LoRA for efficient fine-tuning of the LLM. A fixed CLIP ViT-L vision encoder with a resolution of 560$\times$560 is employed, along with a grounded vision encoder initialized from SAM2 weights. The trainable components of the architecture include a pixel decoder ($\mathcal{D}$), LoRA parameters ($\alpha = 8$), a vision projector $\mathcal{P}_v$, and a language projector $\mathcal{P}_t$. For the adaptive image divider, we set the maximum patch number $\mathcal{P}$ to 9 for training. In our training process, we use an effective batch size of 20 over 10 epochs. The learning rate is scheduled to increase linearly to a maximum value of $3 \times 10^{-4}$ over the initial 100 training steps, followed by a gradual decrease governed by a cosine decay strategy. We train GeoPixel on the GeoPixelD dataset for a grounded conversation generation task on two NVIDIA A6000-48GB GPUs, which take around 3 days.

\subsection{Baselines}
To rigorously evaluate the efficacy of the GeoPixel, we introduce three robust baselines for comparative analysis on the GeoPixelD benchmark. The first baseline, LISA$\dagger$, is an improved version of the LISA model, modified to incorporate multitarget segmentation masks within its output pipeline. Furthermore, the tokenizer is updated to include phrase tokens (\verb|<p> and </p>|) essential for the GCG task, allowing precise identification of contextual phrases within descriptive outputs that correspond to the associated segmentation masks.
The second baseline is derived from the PixelLM$\dagger$ model, configured without the SAM encoder. In this setup, the codebook is configured using image feature scaling fixed at a factor of 2, the number of segmentation tokens adjusted to 3, and the vision tower resize parameter defined at 448. Phrase tokens are added, and  \verb|<SEG>| token in data is replaced with multiple codebook tokens according to the selected configuration. The third baseline, GLaMM, specifically focuses on the GLaMM-GCG variant, a model tailored for the Grounded Conversation Generation task. For LISA$\dagger$, PixelLM$\dagger$ and GLaMM-ft model weights are initialized using pretrained LISA-7B-v1, PixelLM-7B and GLaMM-GCG (7B), respectively, and additionally trained on GeoPixelD data for RS-GCG task.
\begin{figure*}[t]
\vskip 0.2in
\begin{center}
\centerline{\includegraphics[width=1\linewidth]{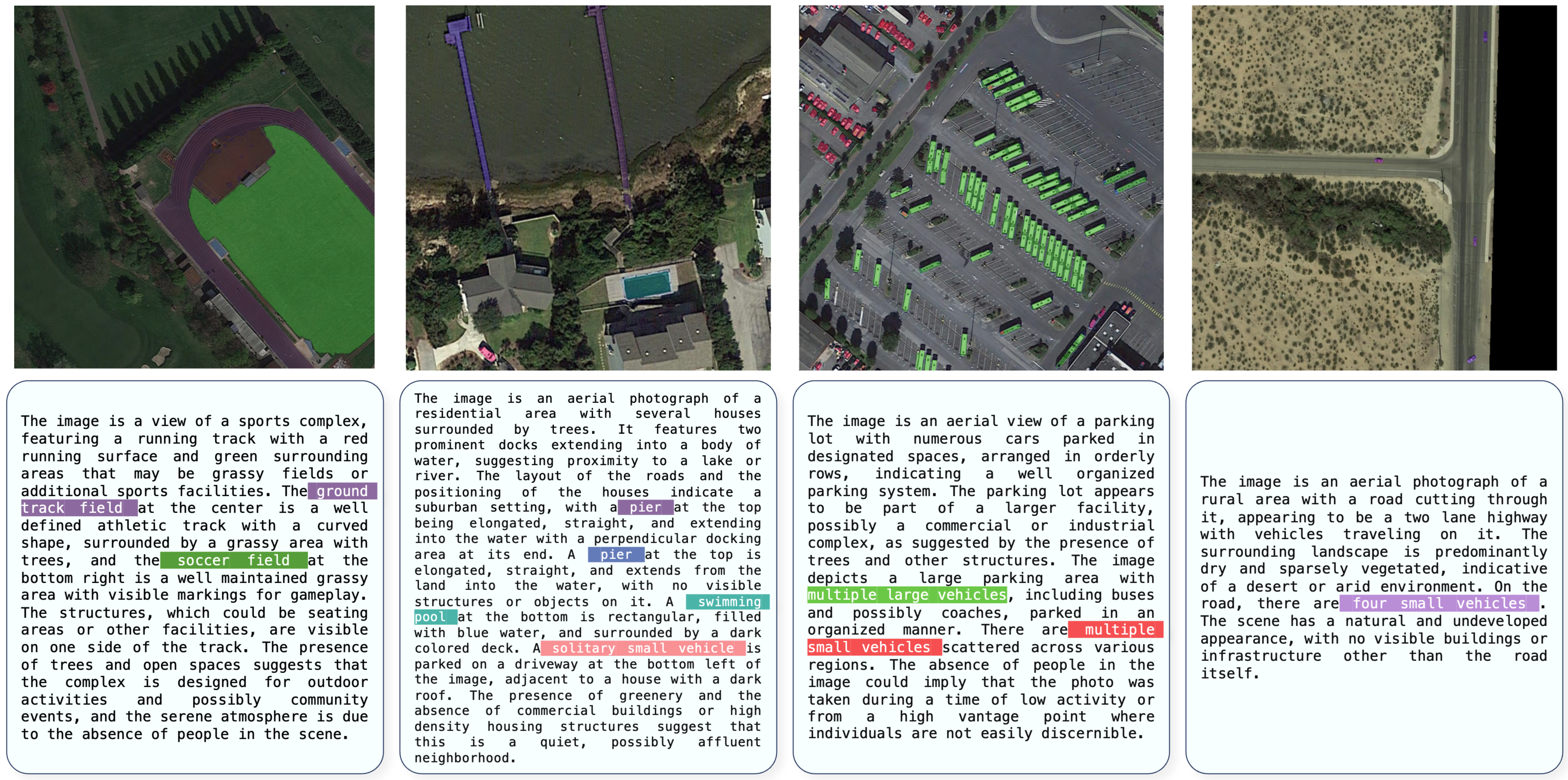}}\vspace{-1em}
\caption{Qualitative results of GeoPixel on RS-GCG. Contextually rich descriptions of RS imagery with grounded object annotations. Depending on object scale and density, it employs instance masks for precise delineation of individual objects (right and middle-right images) while semantic masks capture broader categories, such as large clusters of vehicles or small objects (middle-left and left images).}
\label{fig:geopixeldQR}
\end{center}
\vskip -0.2in
\end{figure*}

\subsection{Results}
\textbf{Remote Sensing Grounded Conversation Generation: }Table~\ref{tab:res} provides a comparative analysis of the performance of various models on the RS-GCG task. The models are evaluated across different metrics, including CIDEr, METEOR, AP50, mIoU, and recall, segmented into Uni-Target, Multi-Target, and Overall categories. GeoPixel demonstrates superior performance in all metrics compared to the baselines showing better fluency and text relevance in textual outputs. In more complex multi-target scenarios, GeoPixel maintains strong performance.
In contrast, LISA$\dagger$ struggles with segmentation-based tasks, as evidenced by its low AP50 scores in all categories. PixelLM$\dagger$ shows a moderate improvement over LISA$\dagger$, benefiting from better image feature scaling and segmentation token adjustments. GLaMM-ft exhibits improved outcomes due to dedicated grounding encoder and GCG pre-training, however, its performance remains inferior to that of GeoPixel. 
Figure~\ref{fig:geopixeldQR} presents the qualitative results.

\begin{table}[!ht]
\caption{Performance Comparison of GeoPixel in Referring Expression Segmentation on RRSIS-D dataset. The segmentation accuracy based on referring expressions is expressed through the Precision at IoU threshold of 0.5 (P@0.5), Overall Intersection-over-Union (oIoU) and Mean Intersection-over-Union (mIoU).}
\setlength{\tabcolsep}{2pt}
\label{tab:res_ref_phr}
\begin{center}
\begin{small}
\begin{sc}
\resizebox{\columnwidth}{!}{%
\begin{tabular}{@{}lcccccc@{}}
\toprule
\textbf{Method}      & \multicolumn{3}{c}{\textbf{Validation Set}} & \multicolumn{3}{c}{\textbf{Test Set}} \\ \cmidrule(lr){2-4} \cmidrule(lr){5-7}
                     & \textbf{P@0.5} & \textbf{oIoU} & \textbf{mIoU} & \textbf{P@0.5} & \textbf{oIoU} & \textbf{mIoU} \\ \midrule
RRN~\cite{rnn}       & 51.09          & 66.53         & 46.06         & 51.07          & 66.43         & 45.64         \\
CSMA~\cite{ye2019cross}       & 55.68          & 69.68         & 48.85         & 55.32          & 69.39         & 48.54         \\
LSCM~\cite{hui2020linguistic}       & 57.12          & 69.28         & 50.36         & 56.02          & 69.05         & 49.92         \\
CMPC~\cite{cmpc}       & 57.93          & 70.15         & 50.41         & 55.83          & 69.22         & 49.24         \\
BRINet~\cite{brinet}     & 58.79          & 70.73         & 51.14         & 56.90          & 69.88         & 49.65         \\
CMPC+~\cite{cmpc_p}      & 59.19          & 70.14         & 51.41         & 57.65          & 68.64         & 50.24         \\
LGCE~\cite{yuan2024rrsis}       & 68.10          & 76.68         & 60.16         & 67.65          & 76.34         & 59.37         \\
LAVT~\cite{LAVT}       & 69.54          & 77.59         & 61.46         & 69.52          & 77.19         & 61.04         \\
RMSIN ~\cite{rmsin} & 74.66 & 78.27 & 65.10  & 74.26   & 77.79  & 64.20   \\
\midrule
Geopixel-ft   & \textbf{80.00}          & \textbf{81.77}         & \textbf{67.99}         & \textbf{83.33}          & \textbf{84.90}         & \textbf{67.30}         \\ \bottomrule
\end{tabular}}
\end{sc}
\end{small}
\end{center}
\vskip -0.1in
\end{table}

\textbf{Referring Remote Sensing Image Segmentation:} 
This task focuses on segmenting specific regions in aerial imagery guided by textual descriptions. The input prompt used is: \texttt{"Could you provide a segmentation mask for \{referring\_expression\} in this image?"} The model generates the response, \texttt{"Sure, it is <SEG>."} where the corresponding embeddings of \verb|<SEG>| token is subsequently decoded to produce the segmentation mask. To address this task, we fine-tune the GeoPixel model on the RRSIS-D~\cite{rmsin} dataset. The resulting GeoPixel-ft model demonstrates superior performance compared to recent approaches, as shown by results on the RRSIS-D test and validation sets in Table \ref{tab:res_ref_phr}. The qualitative results are provided in Figure~\ref{fig:ref_seg_qualitative}.

\begin{table}[htp!]
\caption{Effect of Inference Resolution. Reported metrics show the relationship between resolution and overall performance.
}
\label{tab:abl_res}
\begin{center}
\begin{small}
\begin{sc}
\resizebox{\columnwidth}{!}{%
\setlength{\tabcolsep}{2.5pt} 
\begin{tabular}{cccccccc}
\toprule
\textbf{Training}& \textbf{Inference} &\multirow{2}{*}{\textbf{CIDEr}}& \multirow{2}{*}{\textbf{METEOR}}& \multirow{2}{*}{\textbf{AP50}} & \multirow{2}{*}{\textbf{mIoU}} & \multirow{2}{*}{\textbf{Recall}}\\
\textbf{Patches} & \textbf{Patches} & &  & &  & \\
\midrule
\multirow{3}{*}{$\mathcal{P} = 9$} & $\mathcal{P} = 1 $  & 14.6 & 23.1 & 12.9 & 47.8 & 32.2 \\
& $\mathcal{P} = 4 $  & 17.7 & 23.9  & 16.6 & 51.8 & 37.1  \\
& $\mathcal{P} = 9 $ & \textbf{20.5} & \textbf{24.3} & \textbf{17.6} & \textbf{52.1} & \textbf{37.4} \\
\bottomrule
\end{tabular}%
}
\end{sc}
\end{small}
\end{center}
\vskip -0.1in
\end{table}

\subsection{Ablation Study}

\textbf{Inference Resolution Effect:} Increasing the number of inference patches demonstrates a consistent improvement across all evaluation metrics, reflecting improved model comprehension of visual content (Table \ref{tab:abl_res}). For example, at $\mathcal{P} = 9$, CIDEr increases from 14.6 to 20.5, and METEOR improves from 23.1 to 24.3, indicating improved semantic understanding as the number of image tokens scales up. The moderate gains observed in mAP and mIoU suggest that while high-resolution inference contributes to superior localization accuracy, competitive performance can still be maintained at lower resolutions when the model is pre-trained at higher resolutions. The superior results associated with training with a high patch count ($\mathcal{P} = 9$) underscore the critical role of incorporating fine-grained spatial details during the training phase for generalized feature learning.

\textbf{Annotation Complexity Effect:} GeoPixel adjusts its masking output based on object size and distribution (as seen in Figure~\ref{fig:geopixeldQR}), utilizing instance masks for precise identification of individual objects, while semantic masks are generated to represent broader categories, such as clusters or small objects. In scenarios requiring both granularity and generalization, the model integrates hybrid annotations, blending instance-level and semantic mask representations(as seen in Figure~\ref{fig:RGC}). The effect of this complexity of the annotation is expressed in Table~\ref{tab:abl_ann} with lowest mask recall seen in the case of mixed annotations.

\begin{table}[t] 
\caption{Effect of Annotation Complexity. Avg. Len is the average character length of captions.}
\label{tab:abl_ann}
\vskip 0.1in
\begin{center}
\begin{small}
\begin{sc}
\resizebox{\columnwidth}{!}{%
\setlength{\tabcolsep}{2.5pt}
\begin{tabular}{lcccccc}
\toprule
\textbf{Data}  & \textbf{Objects} & \textbf{Phrases} &\textbf{Avg. Len} & \textbf{mIoU} & \textbf{Recall} \\
\midrule
Instances only & 1,740 & 1,740 & 634 & \textbf{58.4} & \textbf{48.8} \\
Semantic  only & 21,483 & 698 & 518 & 44.1 & 37.7  \\
Mix data & 38,161 & 2,989 & 737 & 50.9 & 33.3 \\
\bottomrule
\end{tabular}
}
\end{sc}
\end{small}
\end{center}
\vskip -0.1in
\end{table}

Remote sensing images often contain visually similar objects with subtle variations in appearance, spatial arrangement, and positional proximity, yet exhibit significant scale variations across different images. This inherent complexity challenges the model's ability to accurately differentiate between object presence, quantity, and the corresponding type of annotation required (e.g., instance level or semantic level). The challenge is particularly evident in the semantic-only category, where the model exhibits the lowest mIoU scores. This indicates two key challenges: the models ability to cover all instances within a category, leading to complete semantic masks, and its ability to group objects under unified semantic mask rather than individual instance identification. The comparatively low mask recall score in mixed data also suggests that the most difficult scenario is to generalize masking decisions effectively in the presence of visually dense objects due to the scale and spatial variability of objects in the image. 

\textbf{Role of Data Complexity:}
In Table ~\ref{tab:abl_dcomp}, we compare the performance of GeoPixel on different data partitions, segregated according to the level of complexity in masking. Set-1A is less complex, with no intra-class segmentation differences. Each instance of a single class is either individually masked or represented using a semantic mask uniformly across the dataset. Set-1B introduces a higher level of complexity where larger instances within the same class are assigned individual instance masks, while smaller objects are grouped under a common semantic mask. For example, two larger boats may be individually described, while all smaller boats in the image could be grouped together under a single semantic description. This structured ablation helps evaluate how GeoPixel handles varying levels of annotation granularity, providing insights into its ability to generalize across different scales and segmentation strategies. The results indicate that inclusion of more complex annotation (Set-1B) leads to improved performance, especially in terms of segmentation accuracy and descriptive detail, as the model is trained with more diverse mask configurations.

\begin{table}[t]
\caption{Effect of Data Complexity and Training Vision Projection (VP) Layer. T stands for Trainable and F for Frozen. }
\label{tab:abl_dcomp}
\begin{center}
\begin{small}
\begin{sc}
\resizebox{\columnwidth}{!}{%
\setlength{\tabcolsep}{2.5pt} 
\begin{tabular}{ccccccccc}
\toprule
\multicolumn{2}{c}{\textbf{Training Data}} & \multirow{2}{*}{\textbf{VP}} &\multirow{2}{*}{\textbf{CIDEr}}& \multirow{2}{*}{\textbf{METEOR}}& \multirow{2}{*}{\textbf{AP50}} & \multirow{2}{*}{\textbf{mIoU}} & \multirow{2}{*}{\textbf{Recall}}\\
\cmidrule(r){1-2}
\textbf{Set-1A} & \textbf{Set-1B}  & &  & &  & \\
\midrule
$\checkmark$ &  & T & 19.3 & 23.6 & \textbf{18.2} & 48.0 & 33.6 \\
$\checkmark$ &$\checkmark$  & T & \textbf{20.5} & 24.0  & 17.8 & \textbf{51.7} & \textbf{36.7}  \\
$\checkmark$ &$\checkmark$ & F & 18.7 & \textbf{24.4} & 15.3 & 51.6 & 35.1 \\ 

\bottomrule
\end{tabular}%
}
\end{sc}
\end{small}
\end{center}
\vskip -0.2in
\end{table}


\textbf{Vision Projection:} Next we study the effect of training the vision projection layer by comparing the performance when the vision projection layer is fixed or trainable during the fine-tuning stage. Table~\ref{tab:abl_dcomp} summarizes the results. Training the vision projection layer results in an improvement in some metrics, highlighting the role of feature alignment.

\begin{figure}[!ht]
\begin{center}
\centerline{\includegraphics[width=1\linewidth]{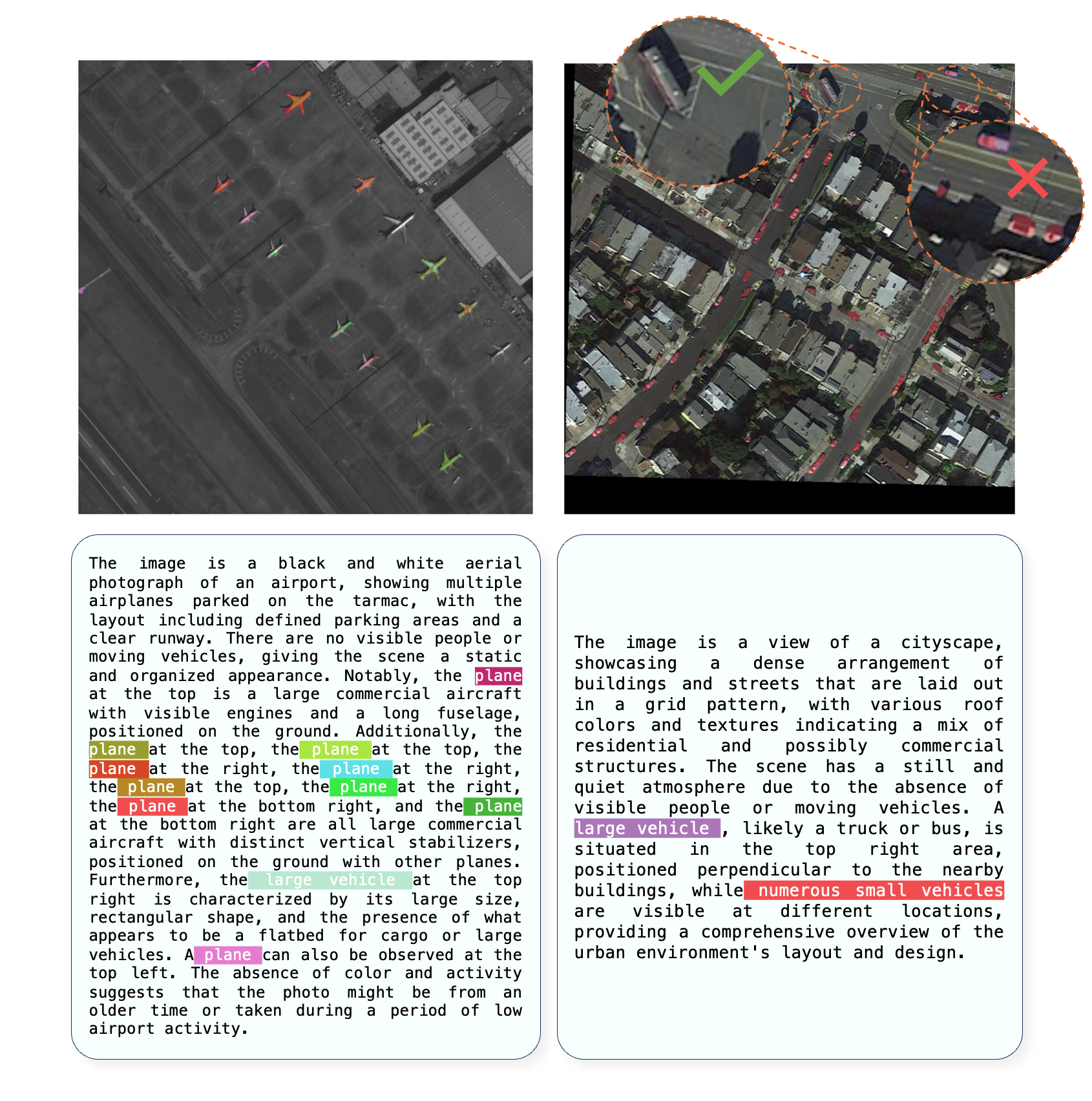}}
\caption{Failure case due to incorrect mask association (left) and wrong instance segmentation in the same spatial region (right).}
\label{fig:fail}
\end{center}
\vskip -0.2in
\end{figure}

\subsection{Limitations and Challenges}
\label{Limitations and Challenges}
While GeoPixel has demonstrated significant advances in pixel-level grounding for high-resolution RS images, several challenges remain. These challenges are particularly evident in the following failure cases (illustrated in Figure~\ref{fig:fail}). The model occasionally produces erroneous masks due to ambiguities in the masking strategy, particularly in determining object presence and quantity, as well as deciding whether semantic segmentation or instance-level annotation is appropriate. An incorrect decision in this regard can result in repetitive descriptions of visually similar objects, leading to inconsistencies in the generated output. Furthermore, such errors may manifest as fragmented or overlapping masks, introducing confusion in object delineation and undermining the overall segmentation quality. Moreover, the model often confuses instance masks within the same spatial location, particularly in densely populated or crowded images.

Future work may focus on addressing these challenges by incorporating more robust masking strategies and dynamic resolution adjustment techniques to improve segmentation accuracy in complex scenes. Additionally, extending GeoPixel's capabilities to integrate multimodal data, such as Synthetic Aperture Radar (SAR) or infrared imagery, could significantly enhance its ability to analyze diverse remote sensing datasets. GeoPixel is a significant step forward in leveraging the potential of LMMs for remote sensing, opening new avenues for research in this critical domain.

\section{Conclusion}
\label{sec:conclusion}

We present GeoPixel, a large multimodal model (LMM) designed specifically for the unique challenges of high-resolution remote sensing (RS) image analysis. GeoPixel introduces a robust end-to-end architecture capable of adaptive image partitioning and pixel-level grounding, enabling the precise interpretation and generation of geospatially aware descriptions in RS imagery. By addressing key limitations of current LMMs, such as low-resolution constraints and coarse object-grounding, GeoPixel provides a fine-grained visual understanding that bridges the gap between language and high-resolution RS data.

\nocite{langley00}

\bibliography{example_paper}
\bibliographystyle{icml2025}

\newpage
\appendix
\onecolumn
\section{GeoPixelD dataset}
\label{Appendix_A}
\textbf{Preprocessing and Marking:}
We utilize the instance-level annotated dataset, iSAID~\cite{waqas2019isaid}, to generate grounded conversations through our annotation pipelines. The images undergo a preprocessing step in which they are cropped into 800 x 800 pixel patches. Objects for instance annotations are selected based on an area threshold to ensure their reasonable size, therefore preventing the marker from obscuring a significant portion of the object and maintaining its distinguishability. A 14 x 14 pixels fixed size marker is used, regardless of the actual dimensions of the object. However, the marker's placement is determined based on the segmentation mask's area and shape. For large objects, the marker is positioned at the center of the mask if the calculated center falls within the mask boundaries; otherwise, it is adjusted to the nearest point on the object's border. For small objects, the center of the bounding box is aligned with a point on the polygon mask boundary, which typically results in an average marker overlap of 50\% with the object.

In addition, multiple marking techniques were also explored, including bounding boxes, masks, contours, and numerical markers, to determine their impact on model accuracy and object fidelity. Our findings reveal that bounding boxes and contours tend to introduce superfluous visual information that can obscure the fine details of the object. In contrast, simple numerical markers placed directly on the object effectively signal its presence without compromising visual clarity or introducing noise, thereby preserving the integrity of object details for enhanced model performance.

\begin{figure*}[htbp]
  \centering
  \includegraphics[width=1\linewidth]{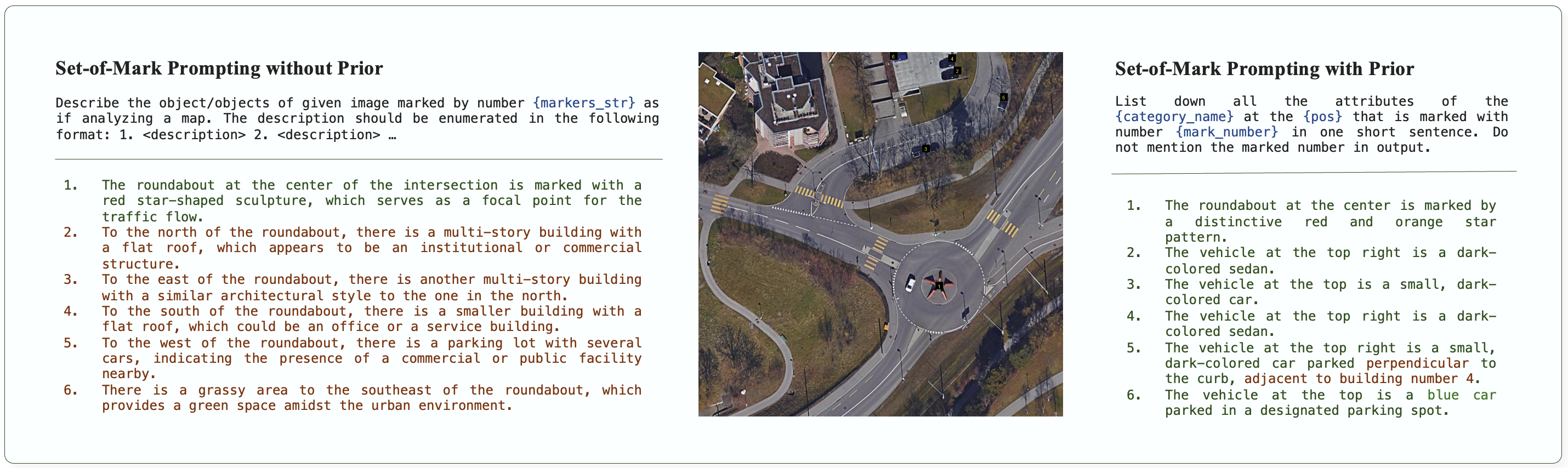} 
  \caption{Comparative effectiveness of SOM prompting methods, highlighting the critical role of priors. Without priors, SOM relies solely on the VLM to detect and describe marked objects independently, resulting in inaccurate descriptions and hallucinated markers in complex remote sensing scenes. In contrast, SOM with priors utilizes explicit marker positions (\{pos\}) and predefined object categories (\{category\_name\}) as priors, providing structured prompts that reduce ambiguity and guide the VLM to produce precise and reliable descriptions. Incorrect parts are noted in \textcolor[rgb]{0.55,0,0}{red} whereas correct parts are noted in \textcolor[rgb]{0,0.39,0}{green}.}
  \label{fig:comp}
\end{figure*}

\begin{figure*}[htbp]
  \centering
  \includegraphics[width=1\linewidth]{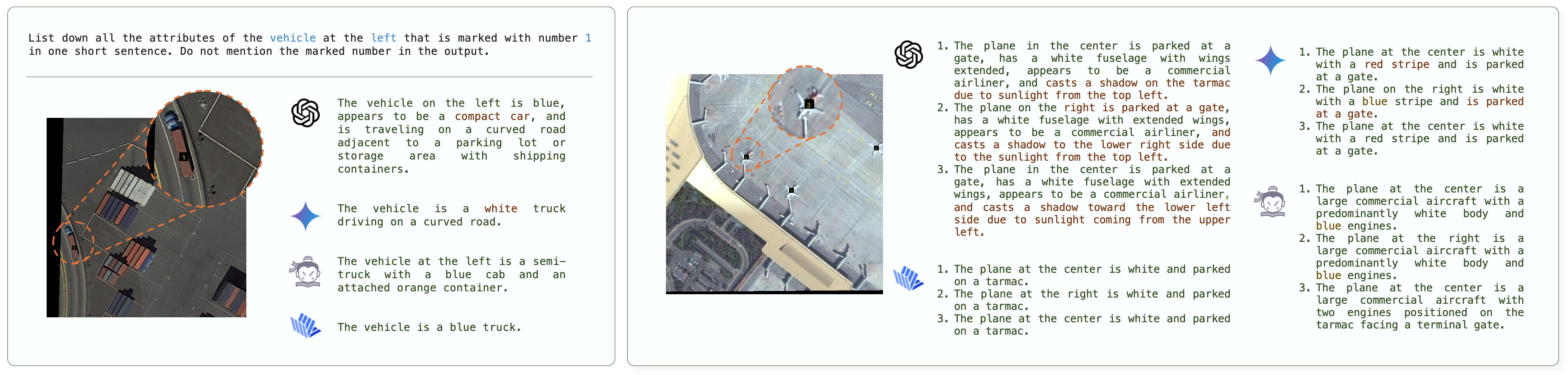} 
  \caption{Comparison of open-source and proprietary models for prior-informed set of marks (SOM) prompting for RS imagery. Incorrect parts are noted in \textcolor[rgb]{0.55,0,0}{red} whereas correct parts are noted in \textcolor[rgb]{0,0.39,0}{green}.}
  \label{fig:compquery}
\end{figure*}

\begin{figure*}[htbp]
  \centering
  \includegraphics[width=1\linewidth]{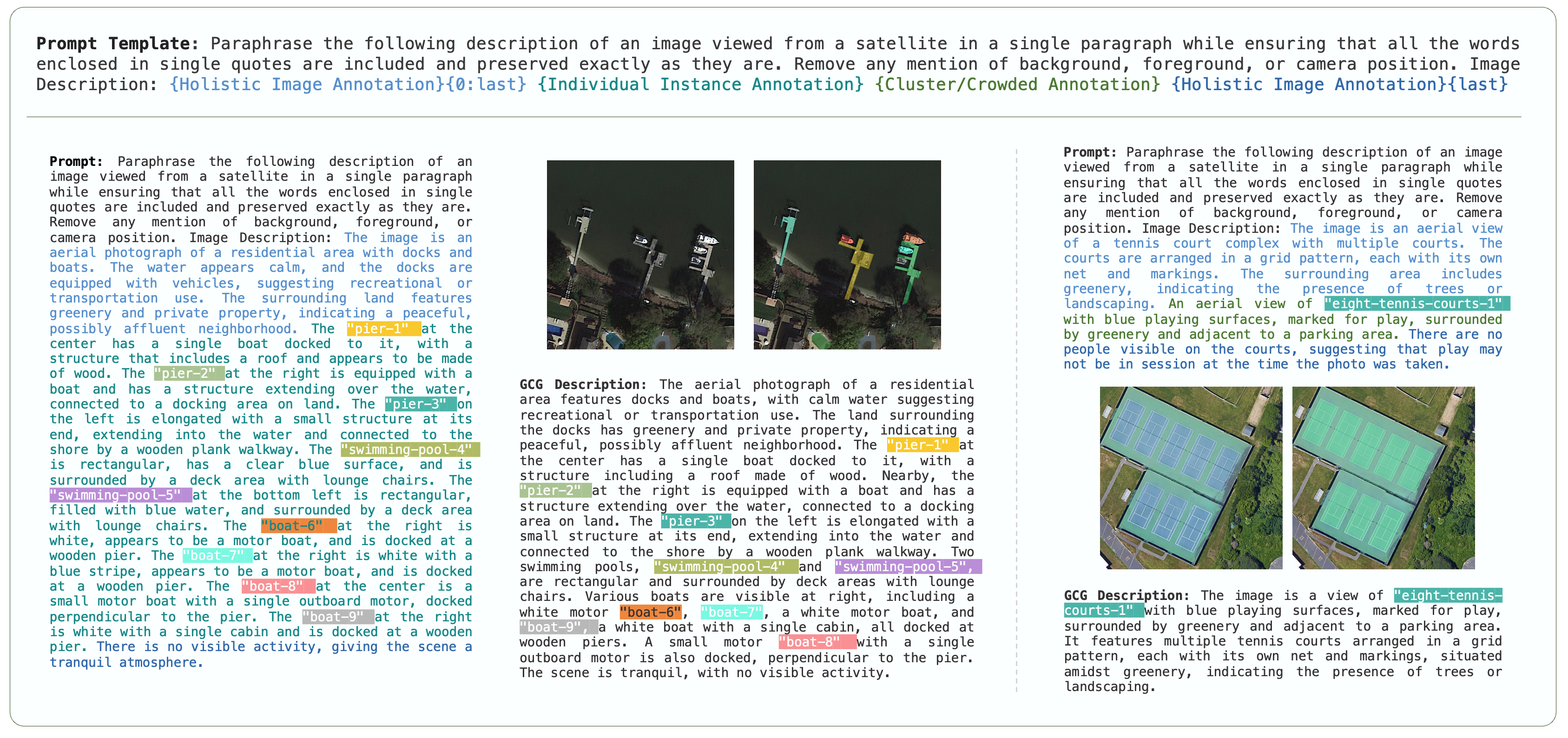} 
  \caption{Unifying Annotations through LLM Paraphrasing and Text Marking to track associated masks. Objects are indexed numerically (e.g., "object-N"), and holistic (blue), individual (teal), and cluster (green) annotations are concatenated into a single image description. Paraphrasing instructions with combined description produce a concise, consistent GCG description that eliminates redundancy while preserving object-mask associations, even with reordering.}
  \label{fig:para}
\end{figure*}

\begin{figure*}[htbp]
  \centering
  \includegraphics[width=1\linewidth]{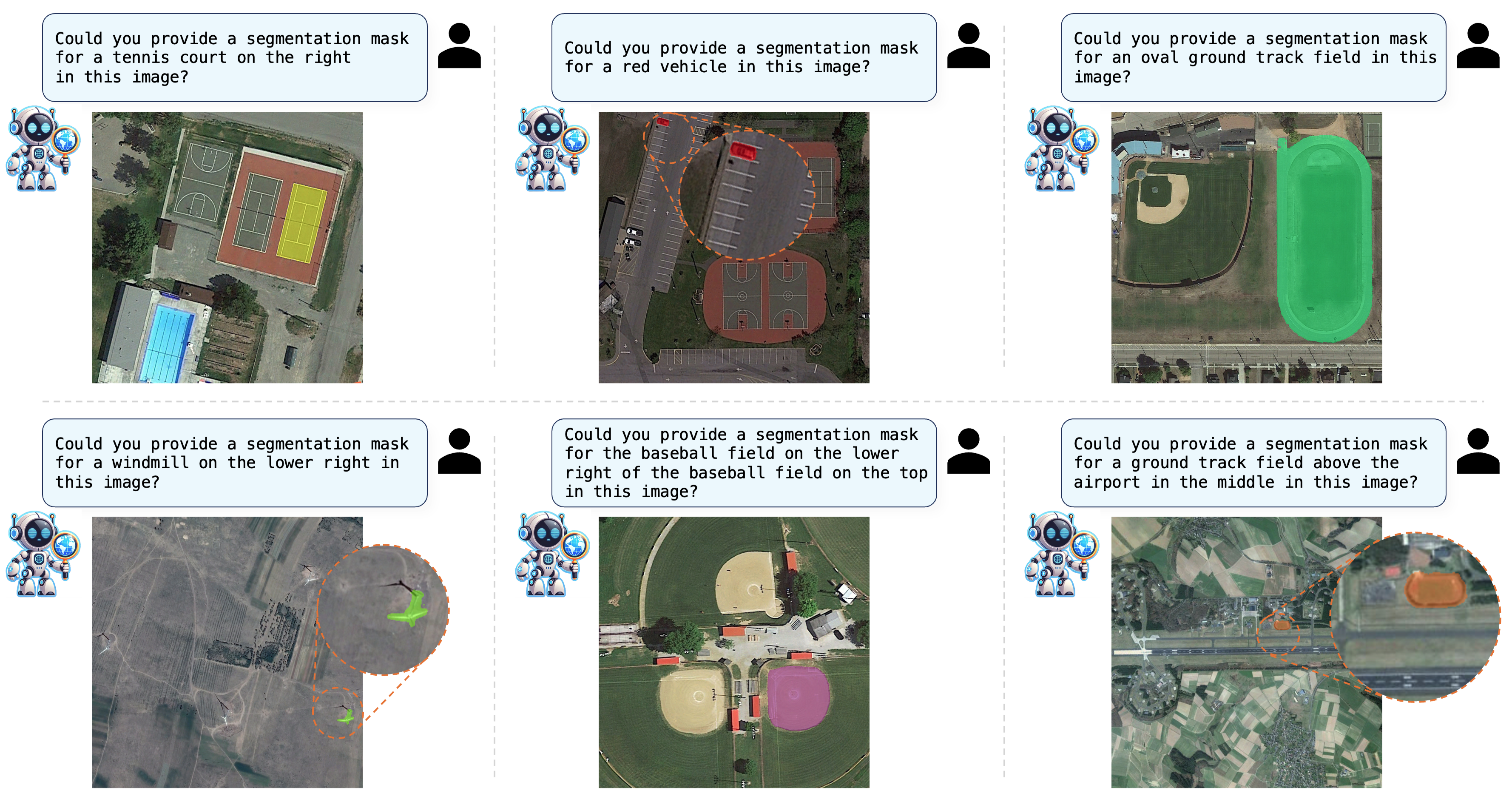} 
  \caption{Qualitative results of GLaMM’s capability in referring remote sensing expression segmentation. The figure highlights Geopixel's ability to interpret referring expressions of varying lengths and generate precise segmentation masks, adapting to scale variations, as shown in the ground track fields. Spatial descriptors (e.g "right", "lower right"), and object characteristics (e.g "red") are interpreted with precision to achieve accurate segmentation. }
  \label{fig:ref_seg_qualitative}
\end{figure*}

\end{document}